\pdfoutput=1

\documentclass[11pt]{article}

\usepackage{acl}

\usepackage{booktabs}
\usepackage{times}
\usepackage{latexsym}
\usepackage{multirow}
\usepackage{graphicx}
\usepackage[T1]{fontenc}

\usepackage[utf8]{inputenc}

\usepackage{microtype}

\usepackage{inconsolata}

\usepackage{times}
\usepackage{soul}
\usepackage{url}
\usepackage{amsthm}
\usepackage{booktabs}
\usepackage{algorithm}
\usepackage{algorithmic}
\usepackage[switch]{lineno}
\usepackage{mathtools,mathrsfs}

\usepackage{enumitem}
\usepackage{arydshln}
\usepackage{subfig}
\usepackage{lipsum}
\usepackage{wrapfig}
\usepackage{textcmds}
\usepackage{amsmath}
\usepackage{amsfonts}
\usepackage{amssymb}

\usepackage{graphicx} 
\usepackage{multirow} 
\usepackage{xcolor}

\def \OURS{MGM}

\newcommand{\blu}{\textcolor{black}}
\title{MGM: Global Understanding of Audience Overlap Graphs\\ for Predicting the Factuality and the Bias of News Media}

\author{
    Muhammad Arslan Manzoor$^{1*}$,
    Ruihong Zeng$^{2*}$,
    Dilshod Azizov$^{1}$, \\
    \textbf{Preslav Nakov$^{1}$ \& Shangsong Liang$^{2\dagger}$} \\
    $^1$Mohamed bin Zayed University of Artificial Intelligence, UAE \\
    $^2$Sun Yat-sen University, China \\
    \texttt{\{muhammad.arslan, preslav.nakov\}@mbzuai.ac.ae}
}

\begin{document}

\maketitle

\begin{abstract}

\def\thefootnote{*}\footnotetext{Equal contribution.}
\def\thefootnote{$\dagger$}\footnotetext{Corresponding author.}
\def\thefootnote{\arabic{footnote}}

In the current era of rapidly growing digital data, evaluating the political bias and factuality of news outlets has become more important for seeking reliable information online. In this work, we study the classification problem of profiling news media from the lens of political bias and factuality. Traditional profiling methods, such as Pre-trained Language Models (PLMs) and Graph Neural Networks (GNNs) have shown promising results, but they face notable challenges. PLMs focus solely on textual features, causing them to overlook the complex relationships between entities, while GNNs often struggle with media graphs containing disconnected components and insufficient labels. To address these limitations, we propose MediaGraphMind (MGM), an effective solution within a variational Expectation-Maximization (EM) framework. Instead of relying on limited neighboring nodes, MGM leverages features, structural patterns, and label information from globally similar nodes. Such a framework not only enables GNNs to capture long-range dependencies for learning expressive node representations but also enhances PLMs by integrating structural information and therefore improving the performance of both models. The extensive experiments demonstrate the effectiveness of the proposed framework and achieve new state-of-the-art results. Further, we share our repository\footnote{\url{https://github.com/marslanm/MGM_code}} which contains the dataset, code, and documentation. 


\end{abstract}

\section{Introduction}

The rise of the Internet has offered many opportunities to publish information and to express opinions~\cite{mehta2023interactively}. Concurrently, this easy means of distribution has accelerated the spread of misinformation and disinformation online which
resembles news in form but lacks the journalistic standards that ensure its quality~\cite{fairbanks2018credibility}. \citet{vosoughi2018spread} has found that
``fake news'' spreads six times faster and reaches much farther than real news. Any delay in profiling in rapidly evolving digital landscapes can lead to unchecked distribution of misleading content \cite{liu2022politics}. 
Profiling news outlets through NLP pipelines offers a proactive approach by enabling the early detection of potentially unreliable sources as soon as they publish content. Since outlets with a history of biased or false information are more likely to do so again, profiling the media in advance allows us to quickly identify probable ``fake news'' by evaluating the reliability of the source itself. \cite{Nakov2024ASO}. 

Early studies on automatic media profiling relied solely on text characteristics~\cite{battaglia2018relational, perez2017automatic}, which has proven particularly challenging. The complexity increases when the text features contain indeterminate noise, leading to classification errors~\cite{baly2018predicting, baly2020written}. Moreover, traditional methods struggle to capture the intricate relationships between entities, such as media outlets, content they publish, and audiences. \citet{EMNLP2022:GREENER} constructed media graphs: nodes represent media, and edges represent audience overlap between media. They proposed a framework that captures both inherent and implicit information about media through interactive learning within the media ecosystem, addressing the limitations of relying solely on textual features. 


We analyze these media graphs and identify two key challenges: disconnected components and label sparsity. Disconnected components prevent GNNs from capturing long-range dependencies, limiting their ability to learn expressive node representations for classification tasks~\cite{longa2024explaining, zhang2024linear}. Prior studies~\cite{yin2024dynamic,tang2024xgnn} address similar issues by using memory-based approaches that store \textit{global information} 
throughout the graph using external memory modules. However, these methods require significant memory to store all node embeddings.

To tackle these challenges, we present \OURS{}, a novel method based on a variational Expectation-Maximization (EM) framework that augments existing Graph Neural Networks (GNNs) to capture and exploit global information in media graphs. \OURS{} seamlessly integrates local and global patterns, node features, and labels from globally similar nodes to enhance performance. Unlike Graph Attention Networks (GATs) \cite{velivckovic2018graph}, which focus solely on local neighborhoods, \OURS{} employs an external memory module to store precomputed node representations of all nodes. This approach not only reduces computational costs \cite{fey2021gnnautoscale} but also facilitates efficient node embedding retrieval. Furthermore, \OURS{} optimizes memory usage by focusing on a small set of candidate nodes, guided by a Dirichlet prior distribution \cite{he2020learning}.


The experimental results show that \OURS{} substantially enhances the performance of baseline GNNs, delivering a 10\% increase across all evaluation measures on the Media Bias/Fact Check (MBFC)\footnote{\href{https://mediabiasfactcheck.com/}{www.mediabiasfactcheck.com}} data feature in the ACL-2020~\cite{baly-etal-2020-written} and the EMNLP-2018~\cite{baly2018predicting} datasets. Despite the lack of rich node features in the media graph, we enhance the dataset by scraping \emph{Articles} and \emph{Wikipedia} descriptions for ACL-2020. Pre-trained language models (PLMs) such as 
BERT~\cite{devlin2018bert}, RoBERTa~\cite{liu2019roberta}, DistilBERT~\cite{sanh2019distilbert}, and DeBERTaV3~\cite{he2021debertav3} 
are fine-tuned to predict political bias and factuality. Where textual data of media are inaccessible, MGM's representation-based probabilities compliment the gap. Moreover, integrating \OURS{}'s probabilities with PLMs enhance the performance for both tasks.
Our contributions are as follows:
    \begin{itemize}

        \item We introduce \OURS{}, an efficient and expressive approach that enhances GNNs for reliable news media profiling by leveraging global information and minimizing memory requirements via a sparse distribution.
        \item We illustrate that \OURS{} consistently outperforms vanilla GNNs for the detection of factuality and political bias across all baselines.
        \item We validate that integrating the \OURS{} features with the PLMs enhances performance and yields state-of-the-art results.
    \end{itemize}


\section{Related Work}
\subsection{Political Bias and Factuality of Media}

Early research on \emph{political bias} detection focused on the analysis of textual content~\cite{afroz2012detecting, battaglia2018relational, perez2017automatic, conroy2015automatic}. To improve the performance, subsequent research added contextual information~\cite{baly-etal-2020-written,hounsel2020identifying,castelo2019topic,fairbanks2018credibility}, including the nuances of multimedia production~\cite{huh2018fighting}, the associated infrastructure~\cite{hounsel2020identifying}, and the social context~\cite{baly-etal-2020-written}. \citet{guo2022measuring} used BERT \cite{devlin-etal-2019-bert} to model the linguistic political bias in news articles. \citet{fan2019plain} used annotated media from \citet{budak2016fair}, analyzing articles for political bias using distant supervision. Various methods measured political bias, including analyzing Twitter interactions \cite{an2012visualizing, ACL2020:Topical:Stance}, often using small datasets only in English \cite{clef-checkthat:2023:task3, clef-checkthat:2023:task4, nakov2023overview, barron2023clef, barron2023overview, azizov2023frank,spinde2022neural}. 

\blu{\citet{lei2022sentence} improved political bias detection through discourse structures, \citet{liu2019detecting} detected frames in gun violence reporting, and \citet{lee2022neus} proposed framework for neutral summaries. \citet{bang2023mitigating} proposed a polarity minimization loss to reduce framing bias in multi-document summarization. \citet{liu2023open} addressed framing bias in event understanding with a neutral event graph induction framework using graph-based approaches. 
\citet{maab2024media} and \citet{lin2024indivec} leveraged LLMs and vector databases for adaptability and explainability. Contributions include frameworks for detecting political bias \cite{trhlik2024quantifying}, media credibility via retrieval-augmented generation (RAG) \cite{schlichtkrull2024generating}, and scalable LLM bias assessment \cite{bang2024measuring}. \citet{demszky2019analyzing} examined polarization on social media, \citet{das2024media} and \citet{zhao2024media} analyzed event relationships in media narratives, and \citet{kameswari2021towards} introduced a corpus quantifying media bias. \citet{kim2024observing} showcased LLMs role in analyzing U.S. cultural patterns and media-driven behaviors.}

The \emph{veracity} of the news media has been explored using PLMs to estimate the reliability of the source, correlated with the ratings of human experts \cite{yang2023large}. \citet{mehta2023interactive} introduced a framework that combines graph-based models, PLMs, and human experience to profile news media, effectively identifying ``fake news'' with minimal human input. 
Recent approaches, such as \citet{baly-etal-2020-written}, used gold labels and various English sources as features to profile media with PLMs. \blu{\citet{azizov2024safari} conducted a cross-lingual evaluation of political bias and factuality.} Although the features of the aforementioned studies are obtained from various sources, they neglect the inherent relationships between the media.

To bridge this gap, graphs emerged as a comprehensive and effective framework for representation learning \cite{mehta2022tackling}. However, this study focused solely on the factuality
task, despite having available political bias labels and generalizing only R-GCN.  \citet{mehta2023interactive} introduced a model that combines graphs, LLMs, and human input for profiling. \citet{EMNLP2022:GREENER} constructed a graph based on the principle of homophily, suggesting that similar media sources attract similar audiences. The framework leveraged the audience overlap of media outlets to build a huge graph that models the interactions between media and to learn expressive representation for the nodes using GNNs. 
However, media graphs are characterized by disconnected components and scarce labels. To overcome these limitations, we propose \OURS{} to effectively capture the information across the entire graph.

\subsection{Graph Neural Networks}
The current design of GNNs follows the message-passing framework \cite{yang2022breaking, chen2024fairness, zeng2024enhancing}, where they learn node representations by aggregating information from local neighbors. However, media graphs suffer from challenges such as multiple disconnected components and limited labels, making it difficult for GNNs to capture long-range dependencies and to learn effective node representations~\cite{longa2024explaining, zhang2024linear}. 
Recent efforts to integrate external memory modules to store the embeddings of all nodes allow GNNs to capture long-range dependencies across graphs~\cite{yin2024dynamic, tang2024xgnn}. In addition, relational GNNs \cite{zhang2024linear} and event relation graphs \cite{lei2022sentence} improve the detection and analysis of political bias.
However, these methods typically require storing embeddings for all nodes in the graph, resulting in high memory costs and low efficiency during testing. 
Unlike previous approaches, MGM focuses on a small set of candidate nodes, which are more likely to be selected as global similar nodes based on a Dirichlet prior distribution applied to the training nodes~\cite{sethuraman1994constructive}. 

\begin{figure*}[!t]
    \centering
  \includegraphics[width=6.45in, height= 1.09in]{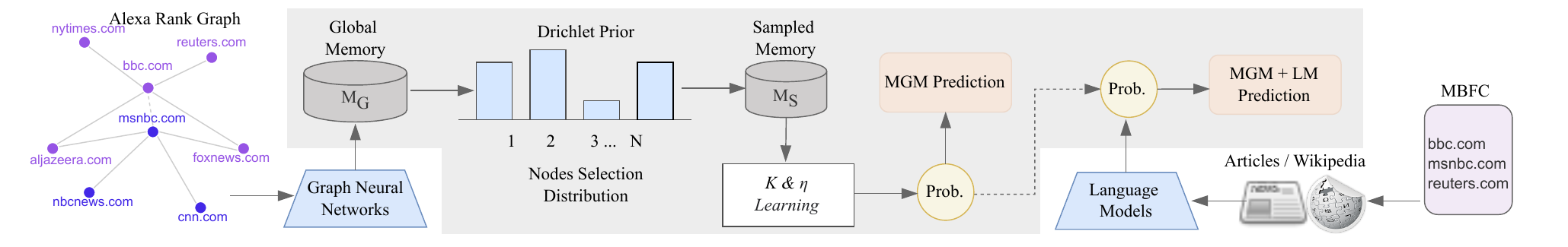}
  \caption{Key components of our proposed approach. The sections highlighted with a grey background represent the architectural contributions introduced by our framework. GNNs store the representation of the media graphs in an external global memory ($M_g$). A Dirichlet prior is used to select the distribution of sparse candidate nodes, which are stored in the sampled memory ($M_s$). The parameters $K$ and $\eta$ control the number of candidate nodes and their influence, balancing local and global information. Since PLMs miss some of the media representation, they leverage MGM representation-based probabilities for the classification task. The detailed pipeline of MGM integration with PLM can be seen in Figure~\ref{method.st} (Appendix~\ref{stages}).}
  \label{fig:main}
\end{figure*}

\section{Methodology}
In this section, we present the problem formulation and provide a detailed description of the proposed framework, which leverages features, structural patterns, and label information from globally similar nodes to enhance GNNs performance. Furthermore, MGM integrates with PLMs to overcome the limitations of existing textual features to detect factuality and political bias.

\subsection{Problem Formulation}
We formulate the news media profiling task as a node classification problem in the semi-supervised graph learning setting~\cite{kipf2016semi,velivckovic2018graph}, where each node represents a news media outlet, the edges capture relationships such as audience overlap, and the node label indicates political bias or factuality, which are available only for a small subset of nodes. Specifically,
let $\mathcal{G} = \{\mathcal{V}, \mathcal{E}, \mathbf{X}, \mathbf{Y}^l \}$ represents a partially-labeled graph, where $\mathcal{V} = \{ v_i \}_{i=1}^{N}$ is a set of nodes, $\mathcal{E}$ is a set of edges, and $N$ is the total number of nodes. The node features are denoted as $\mathbf{X} \in \mathbb{R}^{N \times F}$, where $F$ is the feature dimension. 
Since most nodes are unlabeled, $\mathcal{V}$ can be divided into labeled nodes $\mathcal{V}^l$ with labels $\mathbf{Y}^l$, and unlabeled nodes $\mathcal{V}^u$. The labels $\mathbf{Y}^l \in \mathbb{R}^{N_l \times C}$ are in a one-hot form, where $N_l$ and $C$ represent the number of labeled nodes and the number of classes, respectively.
The goal of semi-supervised learning is to learn the model parameters $\theta$ by maximizing the marginal distribution of the overall labeled nodes, i.e., $p_{\theta}(\mathbf{Y}^l \mid \mathbf{X}, \mathcal{E}) = \prod_{n \in \mathcal{V}^l} p_{\theta} (\mathbf{y}_n \mid \mathbf{X}, \mathcal{E})$ on the training graph. 

\subsection{The MGM Framework} 
\label{model_structure}


Following~\cite{qu2019gmnn, qu2021neural}, we adopt a probabilistic framework for node classification, treating node representations $\mathbf{Z}$ as latent variables determined by a GNN. To improve the performance of the model, we propose to augment the GNNs with information about \textit{global similar nodes}, i.e., nodes in the entire graph that have similar node features and local geometric structures. 
Specifically, we denote the set of global similar nodes of node $n$ as $\mathbf{t}_n \in \{0, 1\}^{N_l}$, where $\mathbf{t}_{nm} = 1$ indicates that node $m$ is a global node similar to $n$. Similarly to node representations, we also regard the similar node indicator $\mathbf{t}_{n}$ as a latent variable. 
Therefore, the joint probability distribution of global information-enhanced method can be factorized as follows:
\begin{align}
    & p_\theta(\mathbf{Y}^l, \mathbf{T}, \mathbf{Z} \mid \mathbf{X}, \mathcal{E})= \\
    &= p_\theta (\mathbf{Z} \mid \mathbf{X}, \mathcal{E}) p_\theta(\mathbf{T} \mid \mathbf{Z}) p_\theta(\mathbf{Y}^l \mid \mathbf{Z}, \mathbf{T}) \, , \nonumber
\end{align}
where $\mathbf{T} = [\mathbf{t}_n]_{n \in \mathcal{V}^l}^\top$ are the global similar nodes of all nodes. 

However, finding global similar nodes with node representations requires computing representations for all nodes, which is expensive in terms of space and time~\cite{fey2021gnnautoscale}. To alleviate this, we propose to store the embeddings of the labeled nodes in the memory and to use them to find global similar nodes. 
Consequently, the distribution of $\mathbf{T}$ can be replaced by $p_\theta(\mathbf{T} \mid \hat{\mathbf{Z}})$, where $\hat{\mathbf{Z}}$ is the embeddings of the labeled nodes in the memory, i.e., $\hat{\mathcal{V}}^l$.
In this case, we can directly retrieve the representation from memory without computing representations for all nodes, thus making it more efficient to obtain the distribution of global similar nodes for both training and prediction. 

To reduce the memory size, we select global similar nodes from a small set of candidate nodes, which are a subset of the training nodes. As a result, only the embeddings of these candidate nodes are stored in memory for prediction.
To achieve this, we assume that $p_{\theta}(\mathbf{T} \mid \hat{\mathbf{Z}})$ is a sparse distribution, concentrated on a few candidate nodes. 
Since the candidate set is not known, we introduce a latent variable $\boldsymbol{\omega}$ for each node $n$, where $\boldsymbol{\omega}_i \in [0,1], s.t. \sum_{i=1}^{N_l} \boldsymbol{\omega}_i = 1$.
Here, $\boldsymbol{\omega}_i$ represents the probability that the $i$-th node in the labeled node is a candidate node.
Inspired by~\cite{he2020learning}, we introduce a prior over $\boldsymbol{\omega}$, i.e. $p_{\alpha}(\boldsymbol{\omega})$ with parameter $\alpha$. This prior is designed to encourage a sparse distribution over $\boldsymbol{\omega}$.
Therefore, the joint distribution of the method is now defined as: 
\begin{align}
    \label{eq:joint_distribution_of_latent_mariables}
    &p_\theta(\mathbf{Y}^l, \mathbf{T}, \mathbf{Z}, \boldsymbol{\omega} \mid \mathbf{X}, \mathcal{E}, \hat{\mathbf{Z}})=p_\alpha(\boldsymbol{\omega}) \\ 
    &p_{\theta}(\mathbf{Z} \mid \mathbf{X}, \mathcal{E}) \ p_{\theta} (\mathbf{T} \mid \boldsymbol{\omega}, \hat{\mathbf{Z}}) \ 
    p_{\theta} (\mathbf{Y}^l \mid \mathbf{T}, \mathbf{Z}) \, . \nonumber
\end{align}
Next, we introduce the parameterization of our probabilistic framework.

\noindent \textbf{Prior distribution over $\boldsymbol{\omega}$.} 
We use the Dirichlet distribution as the prior distribution over $\boldsymbol{\omega}$, i.e., $
p_{\alpha}(\boldsymbol{\omega}) \propto \prod_{i=1}^N \boldsymbol{\omega}_i^{\alpha_i - 1}$, 
where $\alpha_i$ is the concentration parameter of the distribution. The concentration parameter $\alpha$ is a positive value and a smaller value of $\alpha$ prefers a sparser distribution over $\boldsymbol{\omega}$~\cite{he2020learning}. In our experiments, we set $\alpha < 1$ to encourage the sparse nodes distribution.

\noindent \textbf{Prior distribution over node representations $\mathbf{Z}$.}
We model the prior distribution over node representations as Gaussian distributions~\cite{bojchevski2018deep}, 
which are obtained with GNNs due to their effectiveness in graph-learning tasks. 
Therefore, the prior distribution over $\mathbf{Z}$ is defined as follows:
\begin{align}
    p_{\theta} (\mathbf{Z} \mid \mathbf{X}, \mathcal{E}) = \mathcal{N} (\mathbf{Z} \mid \text{GNN}_{\theta}(\mathbf{X}, \mathcal{E}), \sigma^2_1 \mathbf{I}),
\end{align}
where $\sigma^2_1$ is the learned variance of the prior and $\mathrm{GNN}_{\theta}$ is an $L$-layer GNN with parameter $\theta$. 

\noindent \textbf{Prior distribution over $\mathbf{T}$.}
To obtain global similar nodes of node $n$, 
we define a prior distribution over $\mathbf{T}$ as follows: 
\begin{equation}
    p_{\theta} (\mathbf{T} \mid \boldsymbol{\omega}, \hat{\mathbf{Z}}) = \mathrm{Mul}(\mathbf{T} \mid K, f_{\theta}({\boldsymbol{\omega}}, \hat{\mathbf{Z}})), 
\end{equation}
where $\mathrm{Mul}(\cdot)$ represents the multinomial distribution, $K$ denotes the predefined number of global similar nodes, and $f_{\theta}$ is designed as a parameterized function that outputs the 
parameters of the multinomial distribution. 

\noindent \textbf{Prediction of label $\mathbf{Y}$.} 
Finally, we use node representation $\mathbf{Z}$ and information from global similar nodes to predict the label.  Specifically, we leverage the labels of global similar nodes and first predict the label based on its representation:
\begin{align}
    p_{\theta}(\mathbf{Y} \mid \mathbf{Z}) = \mathrm{Cat} (\mathbf{Y} \mid \mathbf{Z} )\, ,
\end{align}
where $p_{\theta}(\mathbf{Y} \mid \mathbf{Z})$
is formulated as a categorical distribution.
Then, we predict the label using the labels of global similar nodes:
\begin{equation}
p_{\theta}(\mathbf{Y} \mid \mathbf{T}) \propto 
\sum_{\mathbf{N, \, M} \in \hat{\mathcal{V}}^l} \mathbf{T}_{\mathbf{NM}} \cdot \mathbf{Y}_\mathbf{M} \, ,
\end{equation}
where $\mathbf{T}_{\mathbf{NM}}$ represents the indices of global similar nodes for the predicted nodes set $\mathbf{N}$. Furthermore, $\mathbf{Y}_\mathbf{M}$ denotes the one-hot labels of the nodes in $\mathbf{M}$, where $\mathbf{M}$ is the set of global similar nodes.
Finally, the predicted label distribution is defined as follows: 
\begin{align}
\label{eq:final_prediction}
p_{\theta} (\mathbf{Y} \mid \mathbf{Z}, \mathbf{T}) &= \eta \, p_{\theta}(\mathbf{Y} \mid \mathbf{Z}) \nonumber \\
&+ (1 - \eta) \, p_{\theta}(\mathbf{Y} \mid \mathbf{T}) \, ,
\end{align}
where $\eta \in [0,1]$ is a trade-off hyper-parameter. When $\eta = 1$, our model only uses local representations of nodes for prediction, which degrades to vanilla GNNs. 
In contrast, when $0 < \eta < 1$, our model predicts the labels of the nodes using information from both local neighbors and global similar nodes.
    
\subsection{Training Process of MGM}
\label{subsec:variational_em}
Next, we explain how to learn the model parameters $\theta$ based on the graph. Ideally, the marginal likelihood should be optimized during training:
\begin{align}
        & p_\theta(\mathbf{Y}^l \mid \mathbf{X}, \mathcal{E}, \hat{\mathbf{Z}})= \\
        &= \int_{\boldsymbol{\omega}} \int_{\mathbf{Z}} \sum_{\mathbf{T}} p_\theta(\mathbf{Y}^l, \mathbf{T}, \mathbf{Z}, \boldsymbol{\omega} \mid \mathbf{X}, \mathcal{E}, \hat{\mathbf{Z}}) \mathrm{d} \mathbf{Z} \mathrm{d} \boldsymbol{\omega} \, .  \nonumber
\end{align}

\begin{table*}[!t]
\centering
\scalebox{0.7}{
\setlength\tabcolsep{12pt}
\begin{tabular}{lcccccc}
\toprule
\multirow{2}{*}{\textbf{Model}} & \multicolumn{3}{c}{\textbf{Fact-2020}} & \multicolumn{3}{c}{\textbf{Bias-2020}} \\
\cmidrule(lr){2-4} \cmidrule(lr){5-7}
& \textbf{Macro-F1} & \textbf{Accuracy} & \textbf{Average Recall} & \textbf{Macro-F1} & \textbf{Accuracy} &  \textbf{Average Recall} \\
\midrule
Majority class & 22.93 $\pm$ 0.00  & 52.43 $\pm$ 0.00 & 33.33 $\pm$ 0.00 & 19.18 $\pm$ 0.00 &  40.39 $\pm$ 0.00 & 33.33 $\pm$ 0.00 \\ \hdashline
GCN & 25.55 $\pm$ 0.94 &	52.55 $\pm$ 0.28 & 34.74 $\pm$ 0.49 & 38.58 $\pm$ 5.13 & 42.90 $\pm$ 4.81  & 41.48 $\pm$ 5.11 \\
\textbf{+ MGM} & \textbf{43.05 $\pm$ 2.03} & \textbf{53.37 $\pm$ 1.00} & \textbf{43.42 $\pm$  1.53} & \textbf{42.77 $\pm$ 1.09} & \textbf{45.23 $\pm$ 1.70} & \textbf{43.80 $\pm$ 3.19} \\ 

GAT & 33.75 $\pm$ 3.12 & 54.18 $\pm$ 0.77 & 39.26 $\pm$ 2.12 & 41.22 $\pm$ 1.79 & 50.34 $\pm$ 0.78 & 48.06 $\pm$ 1.05 \\
\textbf{+ MGM} & \textbf{43.63 $\pm$ 2.80} & \textbf{55.11 $\pm$ 1.44} & \textbf{43.54 $\pm$ 2.71} & \textbf{50.41 $\pm$ 2.86} & \textbf{54.06 $\pm$ 1.98} &  \textbf{51.96 $\pm$ 0.79} \\ 

GraphSAGE & 42.68 $\pm$ 2.55 & 58.02 $\pm$ 1.18 & 45.70 $\pm$ 1.25 & 39.35 $\pm$ 1.07 & 50.00 $\pm$ 1.32 & 49.09 $\pm$ 1.06 \\
\textbf{+ MGM} & \textbf{46.67 $\pm$ 1.58} & \textbf{59.00 $\pm$ 1.00} & \textbf{47.40 $\pm$ 1.67} & \textbf{46.77 $\pm$ 1.82} & \textbf{51.04 $\pm$ 0.67} & \textbf{50.18 $\pm$ 0.92} \\

SGC & 22.73 $\pm$ 0.07 & 51.39 $\pm$ 0.28 & 33.10 $\pm$ 0.18 & 35.37 $\pm$ 0.60 & 45.34 $\pm$ 0.97 & 45.80 $\pm$ 0.76 \\
\textbf{+ MGM} & \textbf{41.28 $\pm$ 1.42}  & \textbf{53.95 $\pm$ 0.77} & \textbf{41.32 $\pm$ 1.22} & \textbf{39.11 $\pm$ 0.51} & \textbf{46.74 $\pm$ 0.78} & \textbf{47.10 $\pm$ 0.74} \\

DNA & 22.75 $\pm$ 0.03 & \textbf{51.74 $\pm$ 0.00}  & 33.33 $\pm$  0.00 & 24.27 $\pm$ 3.02 & 40.69 $\pm$ 0.73 & 35.03 $\pm$ 1.02 \\
\textbf{+ MGM} & \textbf{34.04 $\pm$ 1.60}  & 50.81 $\pm$ 1.30 & \textbf{36.56 $\pm$  1.98} & \textbf{33.22 $\pm$ 1.13} & \textbf{42.55 $\pm$ 2.50} & \textbf{38.59 $\pm$ 1.81} \\ 

FiLM & 43.32 $\pm$ 2.25 & 57.09 ± 0.77  & 44.46 $\pm$ 1.40 & 39.33 $\pm$ 2.76 & 47.55 ± 1.12 & 47.85 $\pm$ 1.07\\
\textbf{+ MGM} & \textbf{49.68 $\pm$ 1.62} & \textbf{57.90 $\pm$ 2.39} & \textbf{49.94 $\pm$ 1.68} & \textbf{45.33 $\pm$ 2.76} & \textbf{48.25 $\pm$ 2.65} &\textbf{48.61 $\pm$ 2.84}\\ 

FAGCN & 24.77 $\pm$ 7.52 & 47.04 $\pm$ 3.71 & 36.12 $\pm$ 5.30 & 19.69 $\pm$ 0.65 & 39.88 $\pm$ 0.28 & 33.71 $\pm$ 0.31\\
\textbf{+ MGM} & \textbf{48.77 $\pm$ 0.00} & \textbf{53.14 $\pm$ 1.66} & \textbf{49.19 $\pm$ 0.00} & \textbf{45.02 $\pm$ 3.00}  & \textbf{45.69 $\pm$ 2.88} & \textbf{45.07 $\pm$ 3.00}\\ 

GATv2 & 51.42 $\pm$ 2.32 & 61.13 ± 1.04  & 55.36 $\pm$ 1.74 & 48.48 $\pm$ 1.68 &  55.11 ± 1.85 & 53.07 $\pm$ 1.75\\
\textbf{+ MGM} & \textbf{54.50 $\pm$ 2.55} & \textbf{62.72 $\pm$ 1.01} & \textbf{57.36 $\pm$ 1.06} & \textbf{52.41 $\pm$ 2.85} & \textbf{55.46 $\pm$ 2.45} & \textbf{54.00 $\pm$ 2.61}\\ 

\bottomrule
\end{tabular}
}
\caption{
Performance of GNN baselines and their \OURS{} enhanced versions for the factuality and political bias tasks on the ACL-2020 dataset, with the majority class baseline and SVM included as na\"{i}ve and non-graphical methods. The higher performance is highlighted in \textbf{bold.}
}

\label{main_results}
\end{table*}

However, the computation of maximizing the marginal likelihood is intractable due to the marginalization of latent variables. As a result, we develop a variational Expectation-Maximization (EM) algorithm \cite{qu2019gmnn} to optimize its evidence lower bound (ELBO) instead:
\begin{align}
\label{eq:elbo}
    & \mathcal{L}_{\mathrm{ELBO}}(\mathbf{Y}^l; \theta, \phi, \alpha, \lambda) = -D_{\mathrm{KL}} (q_{\lambda}(\boldsymbol{\omega}) \, || \, p_{\alpha}(\boldsymbol{\omega})) \nonumber \\ 
    & -D_{\mathrm{KL}} \big[ q_{\phi}(\mathbf{Z} \mid \mathbf{T}, \mathbf{Y}^l) \, || \, p_{\theta}(\mathbf{Z} \mid \mathbf{X}, \mathcal{E})) \big] \notag \\
    & -D_{\mathrm{KL}} \big[ q_{\phi}(\mathbf{T} \mid \mathbf{Y}^l) \, || \, p_{\theta} (\mathbf{T} \mid \boldsymbol{\omega}, \hat{\mathbf{Z}}) \big] \notag  \\
    & +\mathbb{E}_{q_{\phi}(\mathbf{T} \mid \mathbf{Y}^l) q_{\phi}(\mathbf{Z} \mid \mathbf{T}, \mathbf{Y}^l)} \big[ \log p_{\theta}(\mathbf{Y}^l \mid \mathbf{T}, \mathbf{Z}) \big] \, ,
\end{align}
where $D_{\mathrm{KL}}[\cdot || \cdot]$ is the Kullback-Leibler (KL) divergence, $q$ represents the variational distribution to approximate the model posterior distribution and adheres to the following factorization form:\footnote{We omit the dependence of variational distributions on node features $\mathbf{X}$, edges $\mathcal{E}$ and memory $\hat{\mathbf{Z}}$ for brevity.}
\begin{align}
    q_{\lambda}(\boldsymbol{\omega}) q_{\phi}(\mathbf{T}, \mathbf{Z}, \boldsymbol{\omega} \mid \mathbf{Y}^l) q_{\phi} (\mathbf{T} \mid \mathbf{Y}^l) q_{\phi}(\mathbf{Z} \mid \mathbf{T}, \mathbf{Y}^l)  \, , \nonumber 
\end{align}
where $\phi$ and $\lambda$ are variational parameters.

\begin{algorithm}[!t]
\captionsetup{font=small}
\small
\caption{The proposed approach for the node classification task in news media profiling.}
\label{algo:pseudo codes}
\begin{flushleft}
\textbf{Input:} A training graph with labeled nodes $\mathcal{G} = \{\mathcal{V}, \mathcal{E}, \mathbf{X}, \mathbf{Y}^l \}$ and a test graph $\tilde{\mathcal{G}} = \{ \tilde{\mathcal{V}}, \tilde{\mathcal{E}}, \tilde{\mathbf{X}} \}$. \\
\textbf{Output:} Predicted labels $\tilde{\mathbf{Y}}$ for the unlabeled nodes in $\tilde{\mathcal{G}}$.
\end{flushleft}
\begin{algorithmic}[1]
\STATE Pre-train $p_\theta$ according to the message-passing framework.
\WHILE{\textit{no converge}}
    \STATE ${\boxdot}$ \textbf{E-step} 
    \STATE Calculate $q_\phi$ based on $q_{\phi}(\mathbf{T} \mid \mathbf{Y}^l)$ and $q_{\phi}(\mathbf{Z} \mid \mathbf{T}, \mathbf{Y}^l)$.
    \STATE Calculate $q_{\lambda}(\boldsymbol{\omega})$ based on $\prod_{i=1}^{N^l} \boldsymbol{\omega}_i^{{\lambda}_i - 1}$.
    \STATE Update $q_\phi$ and $q_{\lambda}(\boldsymbol{\omega})$ based on Equation~\ref{eq:elbo}.
    \STATE ${\boxdot}$ \textbf{M-step}
    \STATE Calculate $p_\theta$ based on Equation~\ref{eq:joint_distribution_of_latent_mariables}.
    \STATE Update $p_\theta$ based on $p_{\theta}(\mathbf{Y}^l, \mathbf{T}, \mathbf{Z}, \boldsymbol{\omega}\mid \mathbf{X}, \mathcal{E})$ under the distribution $q_\phi$.
\ENDWHILE
\STATE Select the top $M$ nodes that occupy $90\%$ of the probability mass and their corresponding memorized embeddings as $\hat{\mathbf{Z}}_{{\boldsymbol{\omega}}}$.
\STATE Classify each unlabeled node in graph with $p_\theta$ and memory $\hat{\mathbf{Z}}_{{\boldsymbol{\omega}}}$ based on
Equation~\eqref{eq:final_prediction}.
\end{algorithmic}
\end{algorithm}

Note that we use the mean-field assumption to approximate the posterior of $\boldsymbol{\omega}$ to simplify the variational distributions. 
For computational convenience, we assume that the variational distributions of these latent variables have the same distribution form as their prior distributions. Hence, we define the variational distributions of $\boldsymbol{\omega}$, $\mathbf{T}$ and $\mathbf{Z}$ to be Dirichlet, multinomial, and Gaussian distributions, respectively. 

Note that the KL divergence in Equation~\eqref{eq:elbo} has a closed-form solution, and we approximate the expectation using a Monte Carlo method by sampling from the variational distributions.
In variational EM, the variational parameters $\phi$ and the model parameters $\theta$ are learned alternately. In the E-step, we fix $\theta$ and update $\phi$ by minimizing the KL divergence to approximate the true posteriors. In the M-step, we fix $\phi$ and update $\theta$ by maximizing the expected log-likelihood.

\subsection{Prediction Process of MGM} 
\label{subsec:prediction}
After training, we expect to obtain a sparse distribution $q_{\lambda}(\boldsymbol{\omega})$, allowing us to select a subset of the candidate nodes. In this case, we can select candidate nodes over a certain probability threshold, thus reducing the memory size and improving the efficiency for prediction.
Specifically, we calculate the expected value of $q_{\lambda}(\boldsymbol{\omega})$ for each node $i$, which is given by $\mathbb{E}_{q_{\lambda}(\boldsymbol{\omega})} [\boldsymbol{\omega}_i] = {\lambda_i} / {\sum_{j=1}^{N_l} \lambda_j}$, and then we select the top-$M$ nodes that occupy 90\% of the probability mass as candidate nodes.

We then leverage the embeddings of the memorized candidate nodes $\hat{\mathbf{Z}}_{{\boldsymbol{\omega}}}$ and $p_{\theta}$ to predict the labels of the test nodes $\tilde{n}$ based on Equation~\eqref{eq:final_prediction}.
We also provide an overview of the optimization process of the~\OURS{} model for the news media profiling in Algorithm~\ref{algo:pseudo codes}.

\subsection{Enhancing PLMs Predictions with MGM}
Next, we demonstrate how \OURS{} improves the performance of PLMs by incorporating information from global similar nodes.
Given textual features $\mathbf{S}$, such as those from \textit{Articles} and \textit{Wikipedia} pages for the media outlet, we first fine-tune the PLMs using the cross-entropy loss.
Then, we concatenate the predicted label distribution from the PLMs with MGM to obtain the final label distribution:
\begin{align}
    &p_{\psi, \theta}(\mathbf{Y} \mid \mathbf{S}, \mathbf{Z}, \mathbf{T})= \\
    &= \text{Softmax}(\oplus (p_\psi(\mathbf{Y} \mid \mathbf{S}), p_\theta(\mathbf{Y} \mid \mathbf{Z}, \mathbf{T})) \mathbf{W} + \mathbf{b}) \, , \nonumber
\end{align}
where $\psi$ are the parameters of the fine-tuned PLMs, $\oplus$ is the concatenation operation, $p_\psi(\mathbf{Y} \mid \mathbf{S})$ is the label distribution predicted by the fine-tuned PLMs, $p_\theta(\mathbf{Y} \mid \mathbf{Z}, \mathbf{T})$ is the label distribution predicted by MGM, which is based on Equation~\eqref{eq:final_prediction}, $\mathbf{W}$ and $\mathbf{b}$ are the parameters of the linear classifier. More details are given in Figure~\ref{method.st} and Appendix~\ref{stages}. 

\begin{table*}[!t]
    \centering
    \scalebox{0.7}{
    
    \setlength\tabcolsep{4pt}
    \begin{tabular}{@{}lcccccc@{}}
    \toprule
        \multirow{2}{*}{\textbf{Model}} & \multicolumn{2}{c}{\textbf{Fact-2020}} & & \multicolumn{2}{c}{\textbf{Bias-2020}} \\
        \cmidrule(lr){2-3} \cmidrule(lr){5-6}
        & \textbf{Macro-F1} \dag / \S & \textbf{Average Recall} \dag / \S & & \textbf{Macro-F1} \dag / \S & \textbf{Average Recall} \dag / \S \\
        \midrule
        GCN  & 42.04 $\pm$ 1.91 / \textbf{43.05 $\pm$ 2.04} & 41.97 $\pm$ 1.77 / \textbf{43.42 $\pm$ 1.53} & & \textbf{42.77 $\pm$ 1.10} / 42.37 $\pm$ 2.09 & 43.72 $\pm$ 3.15 / \textbf{43.80 $\pm$ 3.19} \\ 
        GAT & 40.63 $\pm$ 3.28 / \textbf{43.63 $\pm$ 2.81} & 40.23 $\pm$ 2.88 / \textbf{43.54 $\pm$ 2.71} & & \textbf{50.41 $\pm$ 2.86} / 47.12 $\pm$ 1.69 & \textbf{51.96 $\pm$ 0.79} / 49.39 $\pm$ 2.02 \\
        GraphSAGE & 45.11 $\pm$ 1.44 / \textbf{46.68 $\pm$ 1.59} & 45.69 $\pm$ 1.42 / \textbf{47.40 $\pm$ 1.67} & & 44.96 $\pm$ 1.72 / \textbf{46.78 $\pm$ 1.83} & 49.78 $\pm$ 0.66  / \textbf{51.04 $\pm$ 0.67} \\
        SGC & \textbf{41.29 $\pm$ 1.42} / 39.65 $\pm$ 2.00 & \textbf{41.32 $\pm$ 1.22} / 40.04 $\pm$ 1.50 & & 38.32 $\pm$ 1.41 / \textbf{39.12 $\pm$ 0.51} & 46.74 $\pm$ 0.56 / \textbf{47.10 $\pm$ 0.74} \\
        DNA & 33.48 $\pm$ 3.69 / \textbf{34.05 $\pm$ 1.60} & 35.99 $\pm$ 2.24 / \textbf{36.56 $\pm$ 1.98} & & \textbf{33.22 $\pm$ 1.13} / 32.25 $\pm$ 4.14 & \textbf{38.59 $\pm$ 1.81} / 36.63 $\pm$ 4.53 \\
        FiLM & 45.12 $\pm$ 3.38 / \textbf{49.68 $\pm$ 1.62} & 45.58 $\pm$ 2.78 / \textbf{49.94 $\pm$ 1.68} & & 43.98 $\pm$ 2.57 / \textbf{45.33 $\pm$ 2.76} & 47.86 $\pm$ 1.46 / \textbf{48.61 $\pm$ 2.84} \\
        
        FAGCN & \textbf{48.77 $\pm$ 0.00} / 46.88 $\pm$ 2.88 & \textbf{49.19 $\pm$ 0.00} / 48.05 $\pm$ 2.65 & & \textbf{45.02 $\pm$ 3.00}  / 44.36 $\pm$ 1.24 & \textbf{45.07 $\pm$ 3.00} / 44.47 $\pm$ 1.36 \\
        
        GATv2 & 54.13 $\pm$ 2.93  / \textbf{54.50 $\pm$ 2.55} & 56.82 $\pm$ 1.94 / \textbf{57.36 $\pm$ 1.06} & & \textbf{52.41 $\pm$ 2.85} / 50.44 $\pm$ 0.95 & \textbf{54.00 $\pm$ 2.61} / 52.02 $\pm$ 1.29 \\

    \bottomrule
    \end{tabular}
    }
    \caption{Summary of the  \OURS{} results detailing the performance variation the use of between using full memory (\dag) and a reduced (90\%) memory allocation (\S) for each GNN.}
    \label{mem_tab}
\end{table*}

\section{Experiments}

\subsection{Research Questions}\label{RQ}
We explore the following research questions \textbf{(RQs)}:

\begin{itemize}[noitemsep]
    \item (\textbf{RQ1}) Can \OURS{} tackle disconnected components and label sparsity in media graphs for factuality and political bias detection tasks?
    \item (\textbf{RQ2}) How do the number of global similar nodes \textit{K} and the trade-off hyper-parameter $\eta$ affect the performance of \OURS{}? 
    \item (\textbf{RQ3}) How does the memory module affect the performance of \OURS{}?
    \item (\textbf{RQ4}) How does \OURS{} elevate the performance of PLMs when faced with the challenge of missing text in \emph{Wikipedia} or \emph{Articles}?
\end{itemize}

\begin{table}[!t]
    \centering
    \resizebox{0.43\textwidth}{!}{
    \begin{tabular}{@{}llll@{}}
        \toprule
        \textbf{Model} & \textbf{60\% labels} & \textbf{80\% labels} & \textbf{100\% labels} \\ \midrule
        GAT & 37.90 $\pm$ 0.41 & 39.22 $\pm$ 0.71 & 33.75 $\pm$ 3.12 \\
        \textbf{+MGM} & \textbf{40.80 $\pm$ 3.83} & \textbf{41.99 $\pm$ 1.44} & \textbf{43.63 $\pm$ 2.80} \\ 
        FiLM & 39.89 $\pm$ 1.69 & 38.59 $\pm$ 4.30 & 43.32 $\pm$ 2.25 \\
        \textbf{+MGM} & \textbf{42.93 $\pm$ 4.00} & \textbf{38.62 $\pm$ 2.73} & \textbf{49.68 $\pm$ 1.62} \\ 
        FAGCN & 24.32 $\pm$ 3.18 & 22.73 $\pm$ 0.00 & 24.77 $\pm$ 7.52 \\
        \textbf{+MGM} & \textbf{34.10 $\pm$ 7.36} & \textbf{39.89 $\pm$ 4.04} & \textbf{48.77 $\pm$ 0.00} \\ 
        GATv2 & 40.54 $\pm$ 1.47 & 42.14 $\pm$ 2.76 & 51.42 $\pm$ 2.32 \\
        \textbf{+MGM} & \textbf{42.98 $\pm$ 2.51} & \textbf{44.35 $\pm$ 2.14} & \textbf{54.50 $\pm$ 2.55} \\ 
        \bottomrule
        
    \end{tabular}
    }
    \caption{The impact of different proportions of training labeled data on the performance (Macro-F1) of MGM for the Fact-2020 task.}
    \label{tab:performance_variations}
\end{table}

\subsection{Dataset} \label{dataset_sec}

The dataset for factuality and political bias of news media introduced by \citet{baly-etal-2020-written} comprises 859 media sources\footnote{\url{https://github.com/ramybaly/News-Media-Reliability}}, their domain names and corresponding gold labels. These labels are sourced from MBFC, a platform supported by independent journalists. Factuality is given on a three-point scale: high, mixed, and low. Political bias is also on a three-point scale: left, center, right. \citet{EMNLP2022:GREENER} used Alexa Rank\footnote{\url{http://www.alexa.com/siteinfo}} to create a graph based on audience overlap, using the 859 media as seed nodes. Media sources that shared the same audience, as determined by Alexa Rank, were connected with an edge.
Alexa Rank returned a maximum of five similar media sources for each medium, which could be part of the initial seed nodes or newly identified media. As depicted in Figure \ref{fig:main}, BBC, MSNBC, and Reuters are listed as media sources in the MBFC dataset. The Alexa tool identified five related media for BBC and MSNBC, with an edge connecting them due to their shared audience overlap. In the resulting graph, the nodes represent the media sources, and the edges represent the percentage of audience overlap between two media. We use these publicly available graph data (the only one of its kind) to train GNNs for the factuality and political bias of the news media. More details are given in Appendix~\ref{sec:dataset_Stat}.

\begin{table}[!t]
\scalebox{0.6}{
\begin{tabular}{lcc}
\toprule
\textbf{Model} & \textbf{Macro-F1} \dag / \S & \textbf{Average Recall} \dag / \S \\
\midrule
GCN & \textbf{47.20 $\pm$ 1.54} / 46.52 $\pm$ 1.52 & \textbf{48.13 $\pm$ 1.19} / 47.60 $\pm$ 1.16 \\
GAT & \textbf{54.99 $\pm$ 4.14} / 53.65 $\pm$ 2.79 & \textbf{57.15 $\pm$ 4.05} / 55.85 $\pm$ 2.27 \\
GraphSage & 46.54 $\pm$ 1.65 / \textbf{47.86 $\pm$ 1.38} & 49.09 $\pm$ 0.87 / \textbf{50.91 $\pm$ 1.07} \\
SGC & 44.60 $\pm$ 2.41 / \textbf{45.16 $\pm$ 2.29} & 45.82 $\pm$ 0.73 / \textbf{46.03 $\pm$ 1.80} \\
DNA & \textbf{34.93 $\pm$ 3.95} / 33.88 $\pm$ 1.52 & \textbf{36.71 $\pm$ 3.83} / 35.35 $\pm$ 1.68 \\
FiLM & 51.06 $\pm$ 2.24 / \textbf{51.47 $\pm$ 2.47} & 51.35 $\pm$ 2.16 / \textbf{52.13 $\pm$ 2.01} \\
\bottomrule
\end{tabular}
}
\caption{Summary of the  \OURS{} results detailing performance variations between using full memory (\dag) and a reduced 90\% memory allocation (\S) for each GNN across Fact-2018 task. The best performance per base model is marked in \textbf{bold}.}
\label{2018_mem_tab}
\end{table}

\subsection{Baselines} \label{baseline}
For evaluation, we consider two categories of baselines, including GNN-based and PLM-based models. For GNN models, we select eight well-known models, including GCN \cite{kipf2016semi}, GraphSAGE \cite{hamilton2017inductive}, GAT \cite{velivckovic2018graph}, SGC \cite{wu2019simplifying}, DNA \cite{fey2019just}, FiLM \cite{brockschmidt2020gnn}, FAGCN \cite{bo2021beyond} and GATv2 \cite{brody2022attentive}. More details on these GNN baselines are provided in the Appendix~\ref{append:baselines}. For PLMs, we use four state-of-the-art encoder models, including BERT, RoBERTa, DistillBERT, and DeBERTaV3. Next, we compare our results with state-of-the-art results for the factuality and political bias of the news media \cite{EMNLP2022:GREENER, mehta2022tackling}.

\begin{table}[!t]
\centering
\resizebox{0.4\textwidth}{!}{
\begin{tabular}{lcccc}
\toprule
\textbf{Model} & \textbf{Macro-F1} & \textbf{Average Recall}  \\
\midrule
Majority class & 22.47 $\pm$ 0.00 & 33.33 $\pm$ 0.00 \\
SVM & 41.78 $\pm$ 0.00 & 48.89 $\pm$ 0.00 \\ \midrule
GCN & 48.63 $\pm$ 2.19 & 48.16 $\pm$ 2.49 \\
\textbf{+ MGM} & \textbf{49.21 $\pm$ 1.54} & \textbf{51.13 $\pm$ 1.19}  \\ 
GAT & 46.63 $\pm$ 3.53 & 52.25 $\pm$ 4.20 \\
\textbf{+ MGM} & \textbf{54.99 $\pm$ 4.14} & \textbf{57.15 $\pm$ 4.05}  \\ 
GraphSAGE & 41.77 $\pm$ 0.22 & 48.65 $\pm$ 0.22 \\
\textbf{+ MGM} & \textbf{47.86 $\pm$ 1.38} & \textbf{50.91 $\pm$ 1.07}  \\ 
SGC & 41.06 $\pm$ 0.35 & 44.91 $\pm$ 0.43  \\
\textbf{+ MGM} & \textbf{45.16 $\pm$ 2.29} & \textbf{46.03 $\pm$ 1.80}  \\ 
DNA & 28.24 $\pm$ 1.23 & 33.26 $\pm$ 1.03 \\
\textbf{+ MGM} & \textbf{34.93 $\pm$ 3.95} & \textbf{36.71 $\pm$ 3.83}  \\ 
FiLM & 46.75 $\pm$ 0.79 & 50.92 $\pm$ 1.36  \\
\textbf{+ MGM} & \textbf{51.47 $\pm$ 2.47} & \textbf{52.13 $\pm$ 2.01}  \\
\bottomrule
\end{tabular}
}
\caption{Performance of GNN baselines and their  \OURS{} enhanced versions on the Fact task of EMNLP-2018, with the majority class baseline and SVM included as naive and non-graphical methods. The highest performance is highlighted in \textbf{bold.}}
\label{2018_res}
\end{table}

\section{Discussion}
\subsection{Overall Performance}
\label{mgm_sec}
To answer~\textbf{RQ1}, we conduct factuality and political bias classification experiments in a semi-supervised setting. The experimental results reported in Table~\ref{main_results} demonstrate that \OURS{} can improve the performance of existing GNNs in almost all cases. 
For example, when applied to the Fact-2020 dataset, \OURS{} improves the Macro-F1 performance of GCN, GAT, SGC, and DNA by 17.5\%, 9.8\%, 18.5\%, and 11.4\%, respectively. 
Similarly, for Bias-2020, we can observe that GNNs equipped with \OURS{} consistently outperform the corresponding base models in all evaluation measures. 

We conducted a series of experiments using different proportions of training labels to assess the performance of \OURS{} as shown in Table~\ref{tab:performance_variations}. The results indicate a clear trend: as we increase the percentage of training labels, the model performance improves significantly compared to the baseline. Due to limited data, using a smaller percentage of training labels results in modest improvements over the baseline, constraining the model's ability to generalize well to unseen data. \OURS{} effectively addresses~\textbf{RQ1} by leveraging global similar nodes in media graphs with disconnected components and label sparsity for the detection of factuality and political bias. Our evaluation extends to Fact-2018 and depicts \OURS{}'s stable performance across different datasets presented in Table \ref{2018_res}. The results show that \OURS{} is able to consistently improve all the baselines. Given the reasons described in Appendix \ref{sec:dataset_Stat}, experiments are not conducted on the political bias task of EMNLP-2018.

\begin{table*}[!t]
    \centering
    \resizebox{1.0\textwidth}{!}{
        \begin{tabular}{{llccccccccccccc}}
             \toprule
             & \multirow{4}{*}{\textbf{Model}} & \multicolumn{6}{c}{\textbf{Fact-2020}} & \multicolumn{6}{c}{\textbf{Bias-2020}} \\
             \cmidrule(lr){3-8} \cmidrule(lr){9-14} 
             &  & \multicolumn{3}{c}{\textbf{Articles}} & \multicolumn{3}{c}{\textbf{Wikipedia}} &  \multicolumn{3}{c}{\textbf{Articles}} & \multicolumn{3}{c}{\textbf{Wikipedia}} \\ 
             \cmidrule(lr){3-5} \cmidrule(lr){6-8} \cmidrule(lr){9-11} \cmidrule(lr){12-14}
             & & \textbf{Macro-F1} & \textbf{Accuracy} & \textbf{Avg Recall} & \textbf{Macro-F1} & \textbf{Accuracy} & \textbf{Avg Recall} & \textbf{Macro-F1} & \textbf{Accuracy} & \textbf{Avg Recall} & \textbf{Macro-F1} & \textbf{Accuracy} & \textbf{Avg Recall} \\
             \midrule
             {\multirow{4}{*}{\textbf{STAGE 1}}} & BERT$_{\text{Base}}$ & \textbf{38.27} & \textbf{63.37} & \textbf{39.65} & 34.64 & 59.30 & 37.98 & \textbf{65.38} & \textbf{68.02} & \textbf{64.01} & 58.70 & 62.79 & 58.69 \\
                                        & RoBERTa$_{\text{Base}}$ & 33.55 & 62.79 & 37.65 & 25.29 & 59.30 & 33.36 & 63.34 & 65.70 & 62.41 & 58.73 & 62.78 & 58.71 \\
                                        & DistilBERT$_{{\text{Base}}}$ & 35.27 & 62.21 & 37.65 & 25.27 & 61.01 & 33.30 & 65.04 & 67.44 & 63.91 & 58.71 & 62.85 & 58.72 \\
                                        & DeBERTaV3$_{{\text{Base}}}$ & 25.28 & 61.02 & 33.35 & \textbf{40.81} & \textbf{61.05} & \textbf{40.75} & 58.72 & 62.80 & 58.69 & \textbf{59.65} & \textbf{63.37} & \textbf{59.15} \\ \midrule
            {\multirow{4}{*}{\textbf{STAGE 2}}} & BERT$_{\text{MGM}_{\text{GATv2}}}$  & \textbf{76.18} & \textbf{81.98} & \textbf{71.86} & 73.69 & 81.40 & 70.92 & 83.74 & 84.30 & 83.88 & \textbf{82.25} & \textbf{82.56} & \textbf{81.40} \\
                                        & RoBERTa$_{\text{MGM}_{\text{GATv2}}}$ & 69.89 & 80.23 & 66.85 & 72.73 & 79.65 & 70.52 & 85.51 & 86.05 & 85.38 & 81.32 & 81.98 & 80.71 \\
                                        & DistilBERT$_{{\text{MGM}_{\text{GATv2}}}}$ & 74.55 & 81.40 & 71.48 & 73.03 & 80.23 & 70.84  & 87.20 & 87.79 & 86.90 & 80.68 & 81.40 & 80.25 \\
                                        & DeBERTaV3$_{{\text{MGM}_{\text{GATv2}}}}$ & 64.87 & 77.91 & 62.97 & \textbf{74.56} & \textbf{81.98} & \textbf{73.20} & \textbf{87.71} & \textbf{88.37} & \textbf{87.70} & 80.26 & 80.81 & 79.28 \\
            \bottomrule
        \end{tabular}
    }
    \caption{\underline{\textbf{Stage 1:}} Performance of logistic regression (meta-learner) on PLM probabilities with missing media attributed as probabilities $(0.0, 0.0, 0.0)$. \underline{\textbf{Stage 2:}} Performance of the logistic regression (meta-learner) on PLMs probabilities + $\text{MGM}_{\text{GATv2}}$ probabilities for missing media for factuality and political bias of the ACL-2020 dataset.}
    \label{table:stage1_and_stage2}
\end{table*}

\begin{figure}[!t]
    \centering
  \includegraphics[width=7.8cm,height=5.5cm]{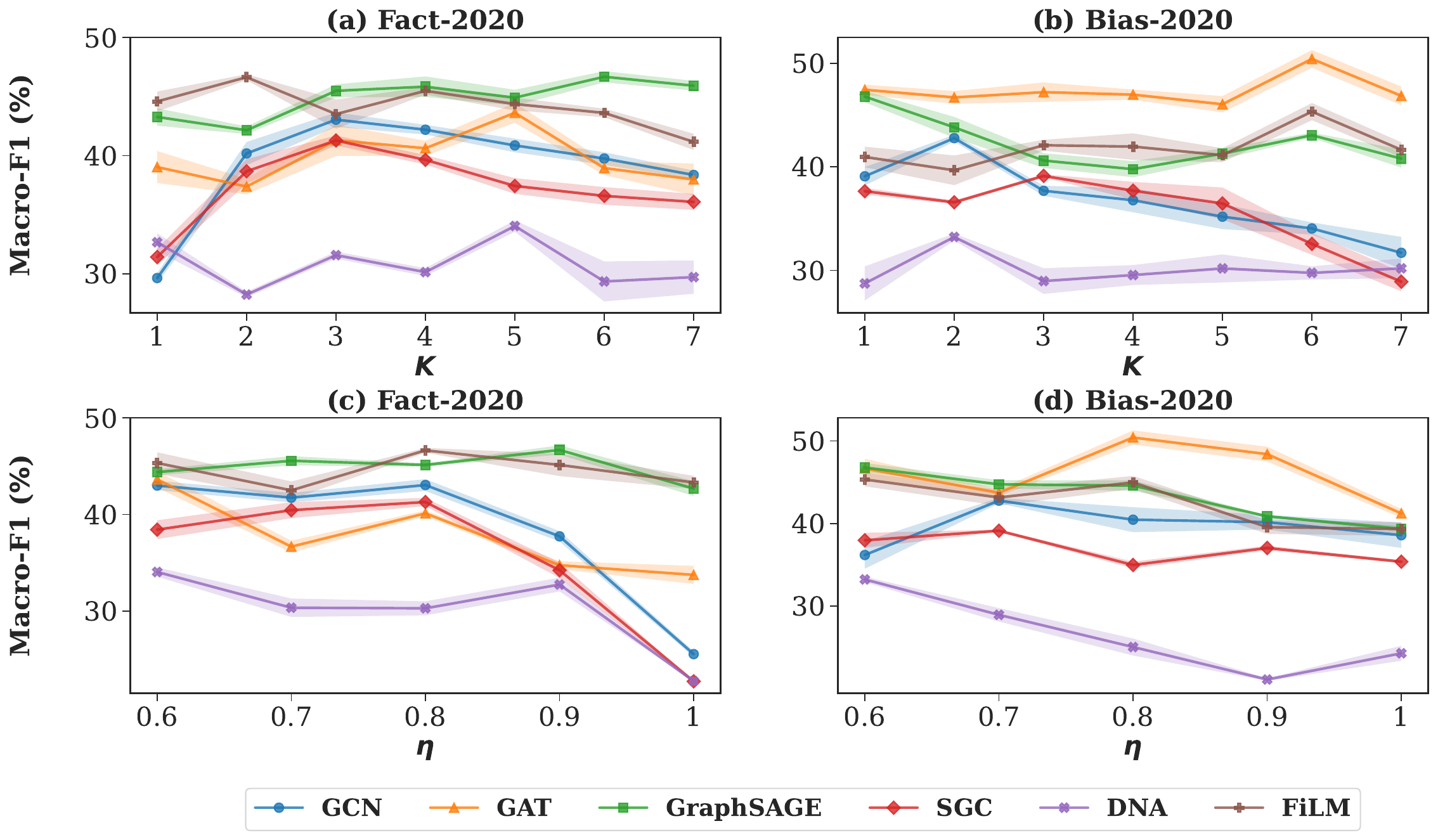}
  \caption{ \OURS{} performance across all GNNs for both tasks, evaluated for different values of $K$ (global similar nodes) and $\eta$ (trade-off hyper-parameter).}
  \label{fig:k_eta}
\end{figure}

\subsection{Impact of the Number of Global Similar Nodes}\label{k_sec}
Next, we turn to \textbf{RQ2} to understand the impact of the number of global similar nodes $K$. Specifically, we investigate the performance of \OURS{} with different values of $K$. As shown in Figures \ref{fig:k_eta}(a) and \ref{fig:k_eta}(b), leveraging a few global similar nodes can improve the performance of the base GNNs. For example, both GCN and SGC exhibit similar patterns, peaking in performance at $K$=3 on the factuality task. The performance of GNNs enhanced with \OURS{} decreases when $K$ exceeds a certain threshold. This is attributed to the introduction of noise by incorporating excessive information from numerous global similar nodes.

\begin{table*}[!t]
    \centering
    \resizebox{1.0\textwidth}{!}{
        \begin{tabular}{{llcccccc}}
             \toprule
             & \multirow{2}{*}{\textbf{Model}} & \multicolumn{3}{c}{\textbf{Fact-2020}} & \multicolumn{3}{c}{\textbf{Bias-2020}} \\
             \cmidrule(lr){3-5} \cmidrule(lr){6-8} 
             & & \textbf{Macro-F1} & \textbf{Accuracy} & \textbf{Avg Recall} & \textbf{Macro-F1} & \textbf{Accuracy} & \textbf{Avg Recall} \\
             \midrule
             & Node classification (NC)~\cite{mehta2022tackling} & 68.90 & 63.72 & - & - & - & - \\
             & Inf\textsc{Op} Best Model~\cite{mehta2022tackling} & 72.55 & 66.89 & - & - & - & - \\
             & GRENNER~\cite{EMNLP2022:GREENER} & 69.61 & 74.27 & - & 91.93 & 92.08 & - \\
             \midrule
             \textbf{STAGE 3} & DeBERTaV3$_{\text{MGM}_{\text{GATv2}}}$ + BERT$_{{\text{MGM}_\text{GATv2}}}$ & 78.43 & 83.04 & 75.03 & 92.64 & 92.98 & 92.67 \\
             \midrule
             {\multirow{3}{*}{\textbf{STAGE 4}}} & DeBERTaV3$_{{\text{MGM}_\text{GATv2}}}$ + BERT$_{{\text{MGM}_\text{GATv2}}}$ + $\text{MGM}_\text{FiLM}$ & \textbf{79.72 $\pm$ 0.00} & \textbf{84.21 $\pm$ 0.00} & \textbf{76.54 $\pm$ 0.00} & 93.04 $\pm$ 0.26 & \textbf{93.45 $\pm$ 0.23} & \textbf{93.19 $\pm$ 0.26} \\
             & DeBERTaV3$_{{\text{MGM}_\text{GATv2}}}$ + BERT$_{{\text{MGM}_\text{GATv2}}}$ + $\text{MGM}_\text{FAGCN}$ & 75.69 $\pm$ 3.49 & 81.29 $\pm$ 3.09 & 72.24 $\pm$ 3.35 & \textbf{93.08 $\pm$ 0.24} & \textbf{93.45 $\pm$ 0.23} & 93.15 $\pm$ 0.34  \\
             & DeBERTaV3$_{{\text{MGM}_\text{GATv2}}}$ + BERT$_{{\text{MGM}_\text{GATv2}}}$ + $\text{MGM}_\text{GATv2}$ & 77.96 $\pm$ 0.30 & 82.69 $\pm$ 0.29 & 74.84 $\pm$ 0.16 & 92.71 $\pm$ 0.46 & 93.10 $\pm$ 0.44 & 92.72 $\pm$ 0.52 \\
            \bottomrule
        \end{tabular}
    }
    \caption{Previous studies \cite{mehta2022tackling, EMNLP2022:GREENER} and our best results. \underline{\textbf{Stage 3:}} We concatenate the probabilities of the best PLMs from \emph{Wikipedia} and \emph{Articles} and use logistic regression to make predictions. \underline{\textbf{Stage 4:}} We use probabilities from the Stage 3 model and concatenate with the probabilities of three GNNs ($\text{MGM}_\text{FiLM}$, $\text{MGM}_\text{FAGCN}$ and $\text{MGM}_\text{GATv2}$).}
    \label{stage34}
\end{table*}

\subsection{Impact of the Trade-off Hyper-Parameter}
\label{eta_sec}
Recall that in Section~\ref{model_structure}, we introduced a hyper-parameter $\eta$ that influences the predicted label distribution.
When $\eta=1$, \OURS{} only relies on local node representations for prediction, degrading to a vanilla GNN. In contrast, when $\eta < 1$, our model incorporates information from both local neighbors and global similar nodes to predict the node labels. 
To further investigate the impact of the trade-off hyper-parameter $\eta$, we analyze the sensitivity of \OURS{} to its value. The experimental results are shown in Figures \ref{fig:k_eta}(c) and \ref{fig:k_eta}(d). We find that compared to $\eta=1$, \OURS{} yields improved performance when $\eta < 1$ in most cases. For example, GCN achieves its best performance when integrated with \OURS{} using an $\eta$ value of 0.8. As a result, the effectiveness of incorporating information from global similar nodes highlighted in the results validates the~\textbf{RQ2}.

\subsection{Effectiveness of the Memory Module}\label{mem_sec}
Recall that in Section~\ref{subsec:prediction}, \OURS{} leveraged a Dirichlet prior to select a small set of candidate nodes and stored their node embeddings in the sampled memory~($\mathbf{M_S}$) for prediction. 
To compare the effectiveness of the sampled memory module to the full memory module~($\mathbf{M_G}$), which stores all the training node embeddings, we conducted a performance comparison between the two memory modules. The experimental results are given in Table~\ref{mem_tab}, and they answer \textbf{RQ3} that \OURS{} using sampled memory achieves a performance comparable to \OURS{} when using full memory. For example, for the GAT model, the performance is higher when using sampled memory compared to when using full memory. This suggests that the sampled memory effectively captures sufficient information, allowing \OURS{} to maintain its performance even with limited memory.
The experimental results on the Fact-2018 dataset reported in Table \ref{2018_mem_tab} also show consistent trends which validate the versatility of the memory module for media graphs.

\subsection{Impact of Integrating MGM with PLMs}

To answer \textbf{RQ4}, we integrate the MGM probabilities with those from deep learning models based on textual features, and we observe that this substantially enhances the performance. Initially, with zero probabilities for missing textual features, we achieved accuracies of 68.02\% for political bias in \emph{Articles}, 63.37\% for \emph{Wikipedia}, 63.37\% for factuality in \emph{Articles}, and 61.05\% for \emph{Wikipedia}, respectively (see Table~\ref{table:stage1_and_stage2}). Replacing the zero probabilities with the best MGM$_{\text{GATv2}}$ improve the performance by up to 30\%. Further concatenating the best model probabilities in stage three led to additional gains, and in stage four, our models outperformed previous state-of-the-art results in political bias and factuality \cite{EMNLP2022:GREENER, mehta2022tackling} (can be seen in Table~\ref{stage34}).

\section{Conclusion \& Future Work}
Our study focused on the underexplored problem of profiling news media in terms of factuality and political bias. To address the shortcomings of existing media graphs, we introduced MediaGraphMind (MGM), an innovative EM framework that significantly enhances the performance of GNNs by leveraging globally similar nodes. The external memory module of \OURS{} efficiently stores and retrieves node representations, addressing the challenge of test-time inefficiency by selecting global similar nodes from a smaller candidate set based on a sparse node selection distribution. Our experiments demonstrate that the integration of \OURS{} features with PLMs consistently improves over existing baselines and establishes a new state-of-the-art results. 

In future work, we plan to explore multi-graph fusion, multi-task learning, and ordinal classification for diverse graph structures in media profiling.

\section*{Limitations}

The graph dataset, originating from the ACL-2020 media nodes, was constructed using the Alexa Rank siteinfo tool, which is currently unavailable. Although the graph aids in the task by capturing the inherent and hidden relationships between media, building such graphs is complex and resource-intensive. The research largely relies on U.S.-centric definitions of political bias (left/center/right), which may not accurately capture the nuanced ideological biases present in news outlets from other cultural or political contexts. Moreover, the available graph is limited to the 2020 dataset. We are actively working on constructing graphs for the latest benchmarks, which include a larger number of media sources and updated MBFC rankings. Moreover, we faced limitations in collecting \emph{Articles} and \emph{Wikipedia} texts from media sources from the ACL-2020 dataset due to the inaccessibility of their websites.

\section*{Ethical Statement}
Optimizing model architectures to enhance energy efficiency in training and inference operations is crucial to reducing environmental impact. Instead of relying on extensive computational resources to train complex models, which significantly increase carbon emissions, we propose improving model performance with less computational power. 
The \emph{Articles} from the news media pages were compiled in strict compliance with legal and ethical standards. We carefully reviewed the terms of use for all websites to ensure that our data collection processes adhered to them. Our compilation focused solely on publicly available data, avoiding paywalls and subscription models. Transparent data collection methods were designed to minimize the impact on source websites, including limiting the access frequency to prevent resource strain. 

\bibliography{anthology,base,custom}

\clearpage
\appendix

\textbf{\large{Appendix}}

\section{GNN Data \& Task Statistics}
\label{sec:dataset_Stat}

Table~\ref{data-details} describes the statistics of the graph data. The factuality is given on a three-point scale: high, mixed, and low. Political bias is also on a three-point scale: left, center, right. \citet{EMNLP2022:GREENER} used the Alexa Rank\footnote{\url{http://www.alexa.com/siteinfo}} (down temporarily) to create a graph based on audience overlap, using 859 media from ACL-2020 \cite{baly-etal-2020-written} as seed nodes. Media sources that shared the same audience, as determined by Alexa, were connected with an edge, provided they met a specific score threshold. Alexa Rank was set to return a maximum of five similar media sources for each medium; these could be part of the initial seed nodes or newly identified media. The primary graph constructed using ACL-2020 dataset media as seed nodes is designated as \emph{level-0}. In this graph, the nodes represent the media sources that publish news or information, and the edges represent the audience overlap for a pair of nodes. The procedure was repeated five times, leading to the formation of five distinct graph levels. With every subsequent iteration, the graph expanded, encompassing media sources previously identified by Alexa Rank. This iterative expansion resulted in a progressive increase in both the number of nodes and edges at each level. \par
Upon analyzing the constructed graphs, we observed several disconnected components, each signifying a unique sub-network of nodes. Naturally, as the graph levels increased, the number of these components decreased. This can be attributed to the fact that an increase in nodes offers more opportunities for components to merge.  We opt for graph level 3 to train GNNs, as detailed in Table \ref{data-details}: it represents the most granular level publicly accessible with fewer disconnected components for both factual and bias tasks. The Alexa Rank tool also generated features for each node in the graph, which we treat as node attributes while training the GNNs. These features include site rank, total sites linked in, bounce rate, and the daily time users spend on the site. These features are the numeric values that are described and normalized in the study \cite{EMNLP2022:GREENER}. We refer to the GNN training tasks as \emph{Fact-2020} and \emph{Bias-2020} for the factuality and political bias tasks, respectively, since both tasks are derived using ACL-2020. As graph-based data becomes increasingly accessible, we focus exclusively on the graph and its inherent features, promoting an approach tailored to such structures. In contrast, \cite{EMNLP2022:GREENER} operates in a supervised setting and uses specialized textual features (e.g., Articles, Wikipedia, Twitter, and YouTube) that are not publicly available. The proposed MGM addresses the unique challenges of the media graph, offering solutions to the research questions described in the designated section \ref{RQ}.

Table \ref{tab:2018-data-details} describes the statistics of the level-3 graph constructed from EMNLP -2018~\cite{EMNLP2022:GREENER} media in the same way explained in Section \ref{dataset_sec}. The EMNLP-2018 dataset comprises 1,066 news outlets, rated on a 3-point scale for factuality (\textit{high, mixed, low}) and a 7-point scale for political bias (\textit{extreme-left, left, center-left, center, center-right, right, and extreme-right}) \cite{baly2018predicting}. A subsequent analysis \cite{baly-etal-2020-written} identified that the labels \textit{center-left} and \textit{center-right} serve as vague intermediate categories, leading to their exclusion. Furthermore, to minimize subjectivity in the annotator decisions, the \textit{extreme-left} and \textit{extreme-right} categories were amalgamated into the \textit{left} and \textit{right} categories, respectively. This adjustment resulted in a simplified 3-point political bias scale (\textit{left, center, right}) and reduced the dataset to 859 outlets as shown in Table \ref{data-details}, published in ACL-2020, which we consider as our main dataset in section \ref{dataset_sec}.

\begin{table}[!t]
\scalebox{0.72}{
\begin{tabular}{ll}
\toprule
\textbf{Property} & \textbf{Specification} \\ \midrule
Nodes & 67,350 \\
Edges & 200,481 \\
Features & 5 \\
Discon. comp. & 44 \\
Avg. nodes / comp. & 1,500 \\
labeled Nodes & 859 (1\%) \\
Unlabeled Nodes & 66,492 (99\%) \\
Tasks & Fact-2020, Bias-2020 \\
Factuality task dist. & high (162), mix (249), low (453) \\
Political Bias task dist. & left (243), center (272), right (349) \\
Training Split & 687 (80\% of 1\%) \\
Test Split & 172 (20\% of 1\%) \\ 
\bottomrule
\end{tabular}
}
\caption{
Statistics about the level-3 graph constructed from ACL-2020 \cite{EMNLP2022:GREENER}.
} 
\label{data-details}
\end{table}

\begin{table}[t!]
\scalebox{0.72}{
\begin{tabular}{ll}
\toprule
\textbf{Property} & \textbf{Specification} \\ \midrule
Nodes &  78429\\
Edges &  232530\\
Features & 5 \\
Discon. comp. & 88 \\
Avg. nodes / comp. & 911 \\
labeled Nodes &  1066 (1.35\%) \\
Unlabeled Nodes &  77363 (98.65\%) \\
Tasks & Fact-2018  \\
Factuality task dist. & high (265), mixed (268), low (542) \\
Training Split & 852 (80\% of 1.35\%) \\
Test Split & 214 (20\% of 1.35\%) \\ 
\bottomrule
\end{tabular}
}
\caption{
Statistics about the level-3 graph constructed from EMNLP-2018.} 
\label{tab:2018-data-details}
\end{table}

\begin{table}[tbh]
\scalebox{0.72}{
\begin{tabular}{ll}
\toprule
\textbf{Property} & \textbf{Specification} \\ \midrule
Tasks & Fact-2020, Bias-2020 \\
Factuality task dist. & high (295), mix (119), low (58) \\
Political Bias task dist. & left (152), center (181), right (139) \\
Training Split & 387  \\
Test Split & 85  \\ 
\bottomrule
\end{tabular}
}
\caption{
Statistics about \emph{Articles} and \emph{Wikipedia} collected from ACL-2020 \cite{EMNLP2022:GREENER} dataset.
} 
\label{data-details_wiki_articles}
\end{table}

\begin{table}[!t]
    \centering
    \resizebox{0.46\textwidth}{!}{
    \begin{tabular}{l|cccc}
        \toprule
        \textbf{Hyper-parameter} & \textbf{BERT} & \textbf{RoBERTa} & \textbf{DistilBERT} & \textbf{DeBERTaV3} \\
        \midrule
        Batch size & 80 & 100 & 120 & 80 \\
        Max length & 512 & 512 & 512 & 512 \\
        Epochs & 3 & 4 & 5 & 5 \\
        Learning rate & 2e-5 & 2e-5 & 2e-5 & 2e-5 \\
        \bottomrule
    \end{tabular}
    }
    \caption{Experimental setup for PLMs.}
    \label{tab:hyper-parameters_lang_model}
\end{table}

\section{Baselines}
\label{append:baselines}
This section summarizes the baseline GNN models that we use as the backbone for our proposed MGM framework to enhance their learning capabilities in the presence of sparsity challenges. \\
\textbf{{GCN}} \cite{kipf2016semi}:
GCN simplifies the convolution operation to alleviate the problem of overfitting and introduces a renormalization trick to solve the vanishing gradient problem. We set the number of hidden neurons to $16$, and the number of layers to $2.$ ReLU \cite{nair2010rectified} is used as the activation function. We do not dropout between GNN layers. 

\textbf{SGC} \cite{wu2019simplifying}:
SGC shows that the graph convolution in GNNs is actually Laplacian smoothing, which smooths the feature matrix so that nearby nodes have similar hidden representations. SGC removes the weight matrices and non-linearity's between layers. In our experiments, we set the number of hidden neurons to $256,$ the number of layers to $2$, and the number of hops at $2.$ We do not dropout between GNN layers.

\textbf{GraphSAGE} \cite{hamilton2017inductive}:
GraphSAGE learns the embeddings of the nodes in the network by sampling and aggregating features from the local neighborhoods of the nodes. GraphSAGE has different variants based on different feature aggregators, and we adopt GraphSAGE with a mean-based aggregator as our baseline. In our experiments, we set the number of hidden neurons at $64,$ and the number of layers to $2.$ ELU \cite{clevert2015fast} is used as the activation function. We do not dropout between GNN layers.

\textbf{GAT} \cite{velivckovic2018graph}:
GAT incorporates the attention mechanism into the propagation step, allowing each node to compute its hidden states by attending to its neighbors using self-attention and multi-head attention strategies. we set the number of hidden neurons to $128$ per attention head and the number of layers to $3.$ The number of heads for each layer is set to $4, 4$ and $6.$ ELU \cite{clevert2015fast}  is used as the activation function. We do not dropout between GNN layers.

\textbf{DNA} \cite{fey2019just}:
DNA uses the jumping knowledge network to enhance the performance of GNNs. This approach enables selective and node-adaptive aggregation of neighboring embeddings, even when they have different localities within the graph. We set the number of hidden neurons to 128, the number of heads to 8, and the number of layers to 4. ReLU \cite{nair2010rectified} is used as an activation function. We set the dropout rate to 0.5 between GNN layers.

\textbf{FiLM} \cite{brockschmidt2020gnn}:
FiLM learns embeddings of nodes in the network by training a linear message function that is conditioned on the features of neighboring nodes. This allows FiLM to effectively capture and incorporate contextual information from neighbors into node embeddings. We set the number of hidden neurons to $320$ and the number of layers to $4.$ We set the dropout rate to $0.1$ between GNN layers.

\textbf{FAGCN}~\cite{bo2021beyond}: FAGCN adopts a self-gating attention mechanism to learn the proportion of low-frequency and high-frequency signals. By adaptively modeling the frequency signals, FAGCN achieves enhanced expressive performance in capturing graph structure and features. We set the number of hidden neurons to $16,$ and the number of layers to $4.$ We set the dropout rate to $0.5$ between GNN layers.

\textbf{GATv2Conv}~\cite{brody2021attentive}: GATv2 introduces a dynamic graph attention variant that reorders internal operations, resulting in a significantly higher level of expressiveness compared to GAT. We set the number of hidden neurons to $64$ per attention head and the number of layers to 3. ELU~\cite{clevert2015fast} is used as an activation function. We do not dropout between GNN layers.

\section{Experimental Settings} 
\label{sec:exp_set}
As mentioned in Section~\ref{subsec:variational_em}, \OURS{} is trained using the variational EM, which 
iteratively maximize the ELBO and the expectation of log-likelihood function through an E-step and an M-step. To optimize the model, we use the Adam optimizer \cite{kingma2015adam} with a learning rate of $0.001.$ The early stopping strategy is implemented with patience in $10$ epochs. In each experiment, we train ~\OURS{} for $50$ iterations to obtain the results.
In order to encourage a sparse node selection distribution, we set the Dirichlet hyper-parameter $\alpha$ to $0.1.$  The hyper-parameter $K$, which determines the number of global similar nodes, is selected from the range $[1, 7]$ through a tuning process. Its value is optimized to achieve the best performance in the validation set for the node classification task. Similarly, the trade-off hyper-parameter $\eta$, which strikes a balance between the utilization of local representations and 
the information from global similar nodes is chosen from the range $[0.6, 1]$ and is tuned to obtain the optimal performance in the validation set for the node classification task. The model is trained for $5$ epochs using different random seeds and mean ± standard deviation is reported. We use the GNN module implementations provided by PyTorch Geometric\footnote{\url{https://github.com/pyg-team/pytorch_geometric/tree/master/examples}} ~\cite{fey2019fast}. 

We optimize hyper-parameters to achieve the best performance on the validation set. In our experiments, we randomly selected 70\% of the dataset as the training set, 10\% as the validation set, and 20\% as the test set. Due to the relatively small size of the training set, we combined the training and validation sets to create a larger final training set. The trade-off hyperparameter eta manages the balance between global and local information that the model considers for the final prediction. Figure 2 shows the Macro-F1 achieved by MGM with different GNNs on the test set at different values of \textit{K} (number of global similar nodes) and eta (trade-off between local and global similar nodes).

\paragraph{Evaluation Measures} We evaluate our frameworks using the mean of three key measures: \emph{Macro-F1, Accuracy,} and \emph{Average Recall.} \emph{Macro-F1} balance precision and recall for each class, ideal for imbalanced datasets. \emph{Accuracy} measures overall correctness, while \emph{Average Recall} highlights the model's sensitivity to different classes. For GNNs experiment, we used an Nvidia 2080 Ti GPU, and for PLMs experiment, we used an NVIDIA A6000 48GB GPU.

\begin{figure*}[!t]
    \centering
    \includegraphics[width=5.56in, height= 2.1in]{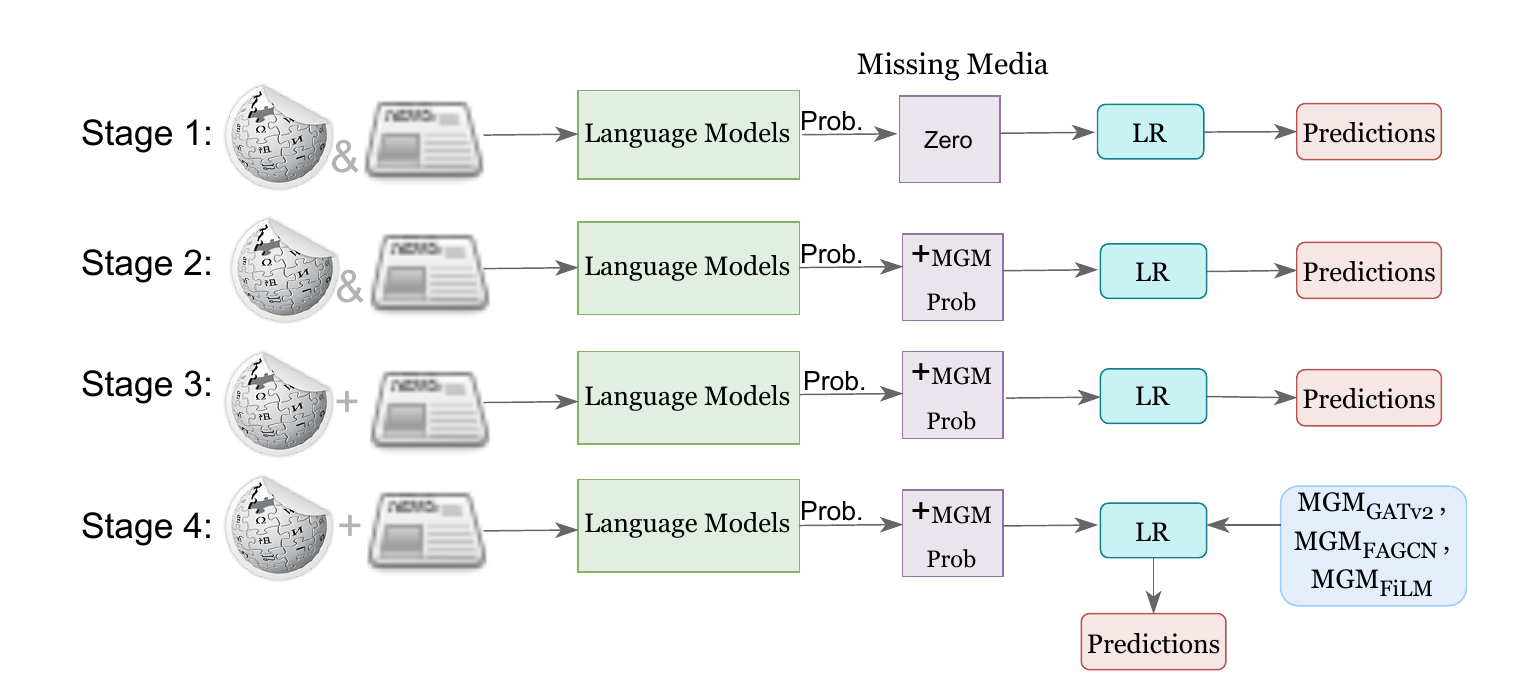}
    \caption{The pipeline of integrating MGM with PLMs. \underline{\textbf{Stage 1:}} We use logistic regression (meta-learner) to make predictions on probabilities obtained from PLMs on 472 media sources. For the remaining media sources, we assign $[0.0, 0.0, 0.0]$ probabilities. \underline{\textbf{Stage 2:}} We use the probabilities produced by PLMs and, for the missing ones, we integrate the probabilities from the best GNN MGM$_{\text{GATv2}}$. The logistic regression is then used to make the predictions. \underline{\textbf{Stage 3:}} We concatenate the probabilities of the best PLM in \emph{Wikipedia} and \emph{Articles} and use logistic regression to make predictions. \underline{\textbf{Stage 4:}} We use the probabilities obtained from Stage 3, which involve concatenating these probabilities with those generated by three GNNs (MGM$_{\text{FiLM}}$, MGM$_{\text{FAGCN}}$, and MGM$_{\text{GATv2}}$)  across five different run seeds. Subsequently, logistic regression is employed to make predictions, and the scores are calculated using the standard deviation.}
    \label{method.st}
\end{figure*}




\section{Collecting Articles \& Wikipedia}
\label{stages}

\emph{Articles.}The article collection involves the following steps: (i) We obtained media sources from the ACL-2020 dataset. (ii) During the article link parsing, we parsed front-page article links from these media sources based on the criteria of selecting only internal links with more than 65 characters and excluding menu button links. (iii) In the article collection stage, we use the selected article links to retrieve the titles and full text of the articles, using scripts and manual testing to ensure effective text extraction, with up to 30 news articles per media. (iv) Finally, the post-processing stage involved formatting the collected data in JSON format. In addition, we specifically targeted sections that focused on political, economic, and social issues sections.

\emph{Wikipedia.} We started by searching for the name of the outlet on the Internet to find the \emph{Wikipedia} link. We ensure that the link leads to a \emph{Wikipedia} page specifically about the media outlet. We then retrieved the text from the \emph{Wikipedia} page using its consistent HTML format. Finally, the post-processing stage involved formatting the collected data in the required JSON format.

In total, from 859 media sources, we have collected data from 472 media sources with \emph{Articles} and \emph{Wikipedia}. Table~\ref{data-details_wiki_articles} provides detailed statistics. Moreover, Figure~\ref{method.st} provides our detailed pipeline for integrating MGM with PLMs.

\section{MGM on Lower Memory Allocation}
\label{lower-mem}

We conduct experiments on 60\% and 80\% and compared with existing results. The results in Table \ref{lw_mem} show that MGM performance produces comparable results even in 80\% of total nodes, but worse in 60\% due to fewer training samples.

\begin{table}[h!]
    \centering
    \resizebox{0.46\textwidth}{!}{
    \begin{tabular}{lcc}
        \toprule
        \textbf{GNN + MGM (Memory \%)} & \textbf{Fact} & \textbf{Bias} \\
        \midrule
        FAGCN + MGM (60\%) & 39.07 & 43.09 \\
        FAGCN + MGM (80\%) & 41.31 & \textbf{46.02} \\
        FAGCN + MGM (90\%) & 46.88 & 44.36 \\
        FAGCN + MGM (100\%) & \textbf{48.77} & 45.02 \\
        \hdashline
        Gatv2 + MGM (60\%) & 45.06 & 45.27 \\
        Gatv2 + MGM (80\%) & 49.77 & \textbf{52.44} \\
        Gatv2 + MGM (90\%) & \textbf{54.51} & 50.44 \\
        Gatv2 + MGM (100\%) & 54.13 & 52.41 \\
        \bottomrule
    \end{tabular}
    }
        \caption{Performance comparison of GNN + MGM at different memory allocations.}
        \label{lw_mem}
\end{table}





\section{MGM Scales to Larger Graphs}
\label{scale-graph}

We primarily focus on addressing the challenges that GNNs face with media graphs containing disconnected components and insufficient labels. However, MGM also shows strong performance when applied to large datasets. We conducted a node classification experiment using the Ogbn-mag \cite{Hu2020OpenGB} dataset, which includes 1,939,743 nodes. The experimental results in Table \ref{graph-scale} show that existing GNNs augmented with MGM can achieve improved performance on large datasets.

\begin{table}[t!]
    \centering
    \begin{tabular}{lc}
        \toprule
        \textbf{Model}      & \textbf{Ogbn-mag}       \\ \midrule
        GraphSAGE           & 46.32 $\pm$ 0.73       \\
        \textbf{+ MGM}  & \textbf{47.94 $\pm$ 0.65}   \\
        \hdashline\noalign{\vskip 0.5ex}
        GAT                 & 44.54 $\pm$ 0.63       \\
        \textbf{+ MGM}  & \textbf{46.28 $\pm$ 0.25} \\
        \hdashline\noalign{\vskip 0.5ex}
        FiLM     &  41.72 $\pm$ 0.22  \\
        \textbf{+ MGM}   & \textbf{43.32 $\pm$ 0.27}   \\
        \hdashline\noalign{\vskip 0.5ex}
        GATv2   & 45.41 $\pm$ 0.42      \\
        \textbf{+ MGM} & \textbf{46.74 $\pm$ 0.36} \\
        \bottomrule
    \end{tabular}
        \caption{Performance comparison on Ogbn-mag.}
        \label{graph-scale}
\end{table}

\section{Training Time of MGM}
\label{MGM_train_time}

We compare the training times of MGM and a vanilla GNN, finding that while MGM requires slightly more training time, the increase is within an acceptable range. In particular, this marginal increase in computational cost is justified by the significant improvement in Macro-F1 scores in Table \ref{train_time}, demonstrating that MGM significantly improves model performance without imposing a considerable training burden.

\begin{table}[h!]
    \centering
    \begin{tabular}{llcc}
        \toprule
        \textbf{Model} & \textbf{Task} & \textbf{Cost Time (m)} & \textbf{Macro-F1} \\
        \midrule
        GATv2 & Fact & 7.41 & 51.42 \\
        + MGM & Fact & 13.04 & \textbf{54.50} \\ \hdashline
        FiLM & Fact & 4.73 & 43.32 \\
        + MGM & Fact & 7.76 & \textbf{49.68} \\ \hdashline
        GATv2 & Bias & 4.50 & 48.48 \\
        + MGM & Bias & 7.97 & \textbf{52.41} \\ \hdashline
        FiLM & Bias & 3.48 & 39.33 \\
        + MGM & Bias & 6.03 & \textbf{45.33} \\ 
        \bottomrule
    \end{tabular}
        \caption{Training Times Comparison between Vanilla GNN and MGM on ACL-2020.}
        \label{train_time}
\end{table}

\end{document}